\documentclass[conference, 10pt]{IEEEtran}
\IEEEoverridecommandlockouts
\usepackage{default}

\def\BibTeX{{\rm B\kern-.05em{\sc i\kern-.025em b}\kern-.08em
    T\kern-.1667em\lower.7ex\hbox{E}\kern-.125emX}}
\begin{document}

\title{Adaptive Block Sparse Regularization \\ under Arbitrary Linear Transform
    \thanks{This work was supported in part by the Japan Society for the Promotion of Science (JSPS) KAKENHI under Grant 23H03419 and Grant 20H04249.}
}

\author{\IEEEauthorblockN{Takanobu Furuhashi}
    \IEEEauthorblockA{
        Nagoya Institute of Technology\\
        Aichi, Japan \\
        clz14116@nitech.jp}
    \and
    \IEEEauthorblockN{Hidekata Hontani}
    \IEEEauthorblockA{
        Nagoya Institute of Technology\\
        Aichi, Japan \\
        hontani.hidekata@nitech.ac.jp}
    \and
    \IEEEauthorblockN{Tatsuya Yokota}
    \IEEEauthorblockA{
        Nagoya Institute of Technology\\
        Aichi, Japan \\
        RIKEN Center for Advanced Intelligence Project\\
        Tokyo, Japan\\
        t.yokota@nitech.ac.jp}
}

\maketitle

\begin{abstract}
    We propose a convex and fast signal reconstruction method for block sparsity under arbitrary linear transform with unknown block structure.
    The proposed method is a generalization of the similar existing method
    and can reconstruct signals with block sparsity under non-invertible transforms, unlike the existing method.
    Our work broadens the scope of block sparse regularization, enabling more versatile and powerful applications across various signal processing domains.
    We derive an iterative algorithm for solving proposed method and provide conditions for its convergence to the optimal solution.
    Numerical experiments demonstrate the effectiveness of the proposed method.
\end{abstract}

\begin{IEEEkeywords}
    Block-sparsity, non-invertible transform, TV regularization, signal recovery
\end{IEEEkeywords}

\section{Introduction}
\zlabel{sec:intro}
Many natural signals exhibit block sparsity.
Block sparsity is a generalization of sparsity that evaluates the sparsity of signal components grouped by block structure.
It has been shown to be more efficient than using mere sparsity in signal reconstruction \cite{baraniukModelBasedCompressiveSensing2010, eldarBlocksparseSignalsUncertainty2010} and is applied in various signal processing and machine learning tasks\cite{yuAudioDenoisingTimeFrequency2008, gribonvalHarmonicDecompositionAudio2003, yuBayesianCompressiveSensing2011,
    wenBlocksparseCNNFast2021}.
Signal recovery based on block sparsity is performed using optimization with an appropriate observation matrix and block structure.

It is generally difficult to know the block structure in advance \cite{kitaharaNonlinearBeamformingBased2022}, so we consider optimizing while adaptively estimating the block structure.
As shown in \zcref{tab:comparison-of-existing-methods}, several methods have been proposed for this purpose \cite{ huangLearningStructuredSparsity2009, fangPatterncoupledSparseBayesian2015, 8319524, jacobGroupLassoOverlap2009, obozinski:inria-00628498, 9729560}.
We focus on Latent Optimally Partitioned $\ell_2$/$\ell_1$ (LOP-$\ell_2$/$\ell_1$), which is superior in both computational efficiency and convexity, and propose a method based on it.
Since non-convex optimization problems have local minima, it is difficult to eliminate the influence of the initial value given to the algorithm.
On the other hand, LOP-$\ell_2$/$\ell_1$ is convex and does not depend on the initial value.
Furthermore, LOP-$\ell_2$/$\ell_1$ is computationally efficient and can be applied to large-scale problems.
\begin{table}[hb]
    \centering
    \caption{
        Comparison of convexity and computational time of existing methods for block sparse regularization when the block structure is unknown.
    }
    \zlabel{tab:comparison-of-existing-methods}
    \begin{tabular}{l|cc}
        \hline
        {Method}                                                                        & {Convexity} & {Computational time}                  \\
        \hline
        {Greedy algorithm\cite{huangLearningStructuredSparsity2009}}                    & \xmark      & $O(N^3)$                              \\
        {Bayesian methods\cite{fangPatterncoupledSparseBayesian2015, 8319524}}          & \xmark      & $O(N^3)$                              \\
        {Latent Group Lasso\cite{jacobGroupLassoOverlap2009, obozinski:inria-00628498}} & \cmark      & Maximum $O(N^3)$ \tablefootnote{
            Depends on the freedom of the assumed block structure.
        }                                                                                                                                     \\
        {\textbf{LOP-$\bm{\ell_2/\ell_1}$}\cite{9729560}}                               & \cmark      & \textbf{$\bm{O(JN)}$} \tablefootnote{
            See \zcref{eq:lop-l2l1-reg} for variable $J$.
        }                                                                                                                                     \\
        \hline
    \end{tabular}
\end{table}

\begin{table}[tb]
    \centering
    \caption{
        Comparison of the application range of existing and proposed methods in terms of the invertibility of linear transform $\bm R$ (see \zcref{eq:aug-lop-l2l1-reg}).
    }
    \zlabel{tab:comparison-of-existing-and-proposed-methods}
    \begin{tabular}{l|cc}
        \hline
        $\bm R$                   & {LOP-$\ell_2$/$\ell_1$ALT} & {LOP-$\ell_2$/$\ell_1$} \\
        \hline
        {invertible}              & \cmark                     & \cmark                  \\
        {\textbf{non-invertible}} & \cmark                     & \xmark                  \\
        \hline
    \end{tabular}
\end{table}

\begin{figure}[t]
    \centering
    \begin{tikzpicture}
        \node (img_true1) {
            \includegraphics[width=0.45\columnwidth]{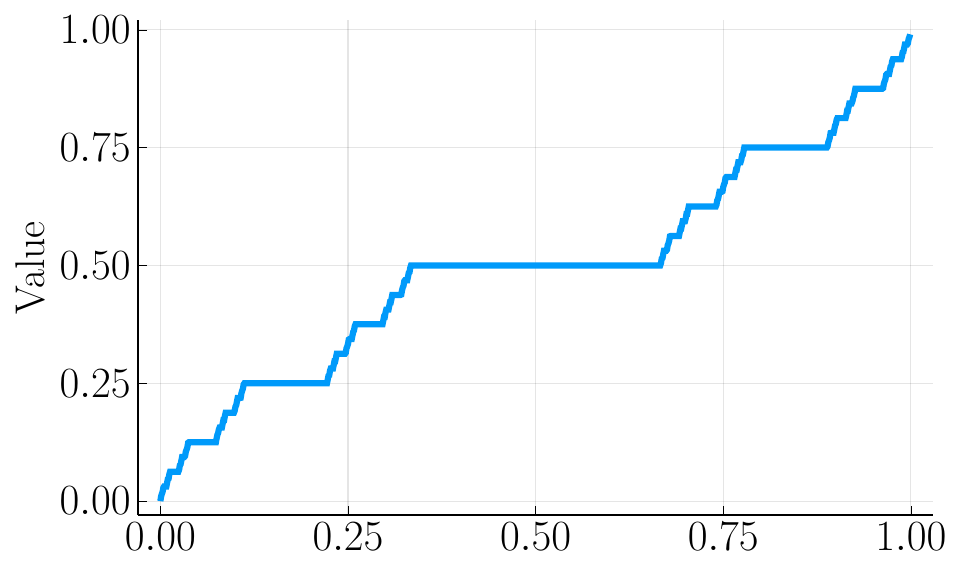}
        };
        \node[above=1.2cm] at (img_true1.center) [xshift=0.2cm] {$\bm x$ (signal space)};
        \node[below=1.2cm] at (img_true1.center) [xshift=0.2cm] {
            \small{Not sparse}
        };
        \node (img_true2) at (4.5, 0) {
            \includegraphics[width=0.45\columnwidth]{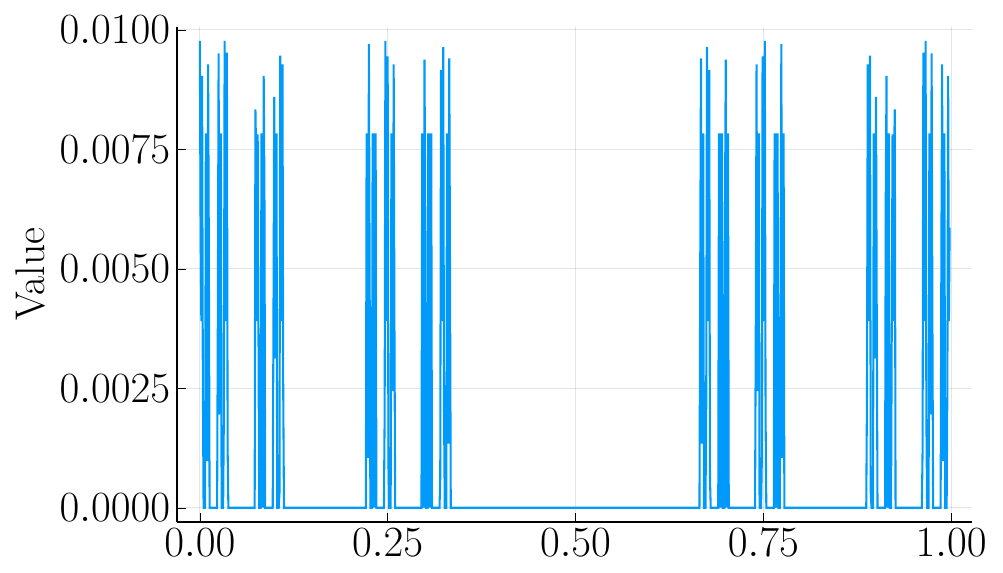}
        };
        \node[above=1.2cm] at (img_true2.center) [xshift=0.2cm] {$\bm{Rx}$ \underline{(feature space)}};
        \node[below=1.2cm] at (img_true2.center) [xshift=0.2cm] {
            \small{\underline{Block sparse}}
        };
        \draw[->,thick] (img_true1) -- (img_true2) node[midway, above=0.1cm] {$\bm R$} node[midway, below=0.2cm] {diff};
    \end{tikzpicture}
    \caption{
        Illustration of block sparsity under non-invertible transform (here, differentiation) in Cantor function \cite{dovgosheyCantorFunction2006}.
    }
    \zlabel{fig:block-sparsity-under-linear-transform}
\end{figure}
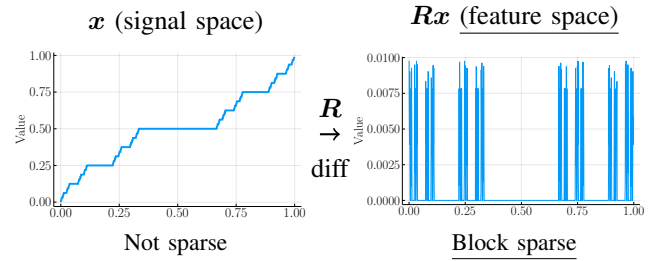

However, LOP-$\ell_2$/$\ell_1$ cannot generally handle block sparsity of a feature vector under non-invertible transforms, as shown in \zcref{tab:comparison-of-existing-and-proposed-methods} and \zcref{sec:limitations-of-lop-l2l1}.
For example, LOP-$\ell_2$/$\ell_1$ can handle block sparsity under invertible transforms such as Fourier transform.
But LOP-$\ell_2$/$\ell_1$ cannot handle block sparsity for differences of feature which is used in TV regularization \cite{rudinNonlinearTotalVariation1992, heTotalVariationRegularizedLowRankMatrix2016}.
TV regularization is a typical approach that introduces a specific linear transformation to existing sparse regularization.
On the other hand, in recent years, low-rank regularization under linear transform has been actively researched.
In \cite{Yokota_2018_CVPR}, low-rank tensor decomposition in the delayed embedded space has been proposed.
In \cite{wangMultiDimensionalVisualData2021, jiangFrameletRepresentationTensor2020} tensor nuclear norm regularization under framelet transformation has been proposed.
In \cite{liGuaranteedMatrixCompletion2019}, low-rank matrix completion under arbitrary linear transformations has been proposed.
This study can be said to be an analogy of these approaches, aiming at adaptive block sparse regularization under linear transformation.

To address this, we propose a novel method called Latent Optimally Partitioned $\ell_2$/$\ell_1$ under Arbitrary Linear Transform (LOP-$\ell_2$/$\ell_1$ALT) regularization, which can evaluate a more general form of block sparsity.
Furthermore, we provide conditions for the convergence of the iterative algorithm for solving the proposed method to the optimal solution.
To demonstrate the effectiveness of the proposed method, we apply it to artificial data and to ion current data obtained using nanopore technology.

\section{Preliminaries}
\zlabel{sec:preliminaries}
\subsection{Estimation of Unknown Block Sparsity}
Let $\bm y = \bm{Lx} + \bm \epsilon \in \C^J$ be the observed vector, $\bm L \in \C^{J \times N}$ be the observation matrix, and $\bm\epsilon \in \C^J$ be the observation error.
Block sparse regularization is performed by adding constraints to the solution of the optimization problem so that the estimated vector $\hat{\bm x} \in \C^N$ or its feature vector $\bm R\hat{\bm{x}} \in \C^K$ becomes block sparse, where $\bm R \in \C^{K \times N}$ is an arbitrary linear transform.
The optimization problem that estimates the vector $\bm x$ while constraining the solution using the LOP-$\ell_2$/$\ell_1$ penalty \zcref{eq:lop-l2l1-penalty} is given as follows:
\begin{equation}
    \min_{\bm x \in \mathbb{C}^N} f(\bm{Lx}) + \lambda\Psi_\alpha(\bm{x}),
    \label{eq:lop-l2l1-reg}
\end{equation}
where $\lambda \in \R_+$ is a regularization coefficient, $f: \C^J \to \R$ is a loss function that encourages $\bm y = \bm{Lx} + \bm{\epsilon}$ to be satisfied, and its proximal operator can be calculated efficiently.
Such functions include squared error $f(\bm{Lx}) = \frac12\norm{\bm y - \bm{Lx}}_2^2$ and absolute error $f(\bm{Lx}) = \norm{\bm y - \bm{Lx}}_1$.

LOP-$\ell_2$/$\ell_1$ \cite{9729560} is a method that performs block-sparse regularization while estimating the unknown block structure using the LOP-$\ell_2$/$\ell_1$ penalty defined as follows:
\begin{equation}
    \Psi_\alpha(\bm x) \coloneqq \min_{\substack{\bm\sigma\in\R^{N} \\
            \norm{\bm{D\sigma}}_1 \leq \alpha}} \varphi(\bm{x}, \bm \sigma),
    \label{eq:lop-l2l1-penalty}
\end{equation}
\zcref{eq:lop-l2l1-penalty} takes small value if $\bm x$ has a block sparse structure with non-overlapping continuous blocks, which is enforced by the inequality constraint $\norm{\bm{D\sigma}}_1 \leq \alpha$ in \zcref{eq:lop-l2l1-penalty}.
In addition, \zcref{eq:lop-l2l1-penalty} is a convex function \cite[Theorem 1]{9729560} and \zcref{eq:lop-l2l1-reg} is a convex optimization problem if $f$ is a convex function such as squared error or absolute error.
$\bm\sigma \in \R^N$ represents latent vectors that indicate the estimated block structure.
In other words, each element $\sigma_n$ of $\bm\sigma$ corresponds to the block that element $x_n$ of $\bm x$ belongs to.
$\alpha \in \R_+$ is a hyperparameter representing the upper limit of the estimated number of blocks.
\zcref{eq:lop-l2l1-penalty} converges to the $\ell_2$ norm when $\alpha $ $\to$ $0$ and to the $\ell_1$ norm when $\alpha$ $\to$ $\infty$.
$\bm D \in \R^{(N-1) \times N}$ is a differential operator for 1D signals represented as follows:
\begin{equation}
    \bm{D} = \begin{bNiceMatrix}
        -1     & 1      & 0      & \Cdots & \Cdots & 0      \\
        0      & \Ddots & \Ddots & \Ddots &        & \Vdots \\
        \Vdots & \Ddots & \Ddots & \Ddots & \Ddots & \Vdots \\
        \Vdots &        & \Ddots & \Ddots & \Ddots & 0      \\
        0      & \Cdots & \Cdots & 0      & -1     & 1
    \end{bNiceMatrix},
    \label{eq:diff_mat_1d}
\end{equation}
and LOP-$\ell_2$/$\ell_1$ can be applied to high-dimensional signals such as images and videos by changing this operator according to the dimension of the signal \cite{9729560}.
The function $\varphi: \C^N \times \R^N \to \R$ is defined as:
\begin{equation}
    \varphi(\bm{x}, \bm \sigma) \coloneqq \sum_{n=1}^N \phi(x_n,\sigma_n),
    \label{eq:varphi}
\end{equation}
where the function $\phi:\C\times\R \to \R_+ \cup \set{\infty}$ is written as:
\begin{equation}
    \phi(x,\tau) \coloneqq \begin{cases}
        \displaystyle \frac{\abs{x}^2}{2\tau} + \frac{\tau}{2}, & \mathrm{if}\ \tau > 0\,;                    \\
        \displaystyle 0,                                        & \mathrm{if}\ x = 0 \text{ and } \tau = 0\,; \\
        \displaystyle \infty,                                   & \text{otherwise.}
    \end{cases}
    \label{eq:var-l2norm}
\end{equation}

\subsection{Limitations of the existing method}
\zlabel{sec:limitations-of-lop-l2l1}
If $\bm R$ is an invertible matrix, i.e., $(\bm R^\top\bm R)^{-1}$ exists, \zcref{eq:lop-l2l1-reg} can be rewritten as follows:
\begin{equation}
    \min_{\bm x \in \mathbb{C}^N} f(\bm{L}(\bm R^\top\bm R)^{-1}\bm R^\top\bm R\bm{x}) + \lambda\Psi_\alpha(\bm{Rx}).
    \label{eq:aug-lop-l2l1-reg-2}
\end{equation}
Then, let $\bm z = \bm R\bm x \in \C^K$ and $\tilde{\bm L} = \bm L(\bm R^\top\bm R)^{-1}\bm R^\top \in \C^{J \times N}$, and \zcref{eq:aug-lop-l2l1-reg-2} can be rewritten as follows:
\begin{equation}
    \min_{\bm z \in \mathbb{C}^K}  f(\tilde{\bm L}\bm z) + \lambda\Psi_\alpha(\bm{z}).
    \label{eq:lop-l2l1-reg-invertible}
\end{equation}
LOP-$\ell_2$/$\ell_1$ can be applied to \zcref{eq:lop-l2l1-reg-invertible} because it has the same form as \zcref{eq:lop-l2l1-reg}.
On the other hand, if $\bm R$ is a non-invertible matrix, \zcref{eq:lop-l2l1-reg} cannot be rewritten in the form of \zcref{eq:lop-l2l1-reg-invertible}, and LOP-$\ell_2$/$\ell_1$ cannot be applied.

\section{Proposed Method}
The proposed LOP-$\ell_2$/$\ell_1$ALT method enables block structure estimation and regularization under various types of linear transforms.
The optimization problem using LOP-$\ell_2$/$\ell_1$ALT penalty is given as:
\begin{equation}
    \min_{\bm x \in \mathbb{C}^N} f(\bm{Lx}) + \lambda\Psi_\alpha(\bm{Rx})
    \label{eq:aug-lop-l2l1-reg}.
\end{equation}

Note that the input of $\Psi_\alpha$ is not $\bm x$ but a feature vector $\bm{Rx}$.
When $\bm R = \bm I$, \zcref{eq:aug-lop-l2l1-reg} is equivalent to \zcref{eq:lop-l2l1-reg}, then the proposed method is an extension of LOP-$\ell_2$/$\ell_1$.
By substituting \zcref{eq:lop-l2l1-penalty}, we can rewrite \zcref{eq:aug-lop-l2l1-reg} as:
\begin{align}
     & \min_{(\bm x, \bm\sigma) \in \C^N \times \R^K} f(\bm{Lx}) + \lambda\varphi(\bm{Rx, \sigma}), \\
     & \quad\text{s.t.}\;\norm{\bm{D\sigma}}_1 \leq \alpha.
    \label{eq:re-aug-lop-l2l1-reg}
\end{align}

Further, we apply the equality constraints $\bm u = \bm{Lx}$, $\bm v = \bm{Rx}$, $\bm \eta = \bm{D\sigma}$ to \zcref{eq:re-aug-lop-l2l1-reg} and replace the inequality constraint $\norm{\bm{D\sigma}}_1 \leq \alpha$ with the indicator function of the $\ell_1$-norm ball:
\begin{equation}
    \iota_{B_1^{\alpha}}(\bm\eta) \coloneqq \begin{cases}
        0,      & \mathrm{if}\ \norm{\bm\eta}_1 \leq \alpha\,; \\
        \infty, & \text{otherwise}.
    \end{cases}
    \label{eq:l1-ball-indicator-function}
\end{equation}
This yields the following expression for \zcref{eq:re2-aug-lop-l2l1-reg}:
\begin{align}
     & \min_{\substack{(\bm x, \bm\sigma) \in \C^N \times \R^K                                                                   \\
            (\bm u, \bm v, \bm \eta) \in \C^J \times \C^K \times \R^{K-1}}}
    \underset{= G(\bm w)}{\underbrace{f(\bm{\bm u}) + \lambda\varphi(\bm{\bm v}, \bm{\sigma}) + \iota_{B_1^{\alpha}}(\bm\eta)}}, \\
     & \quad\footnotesize{
        \text{s.t.}\;
        \underset{= \bm H}{
            \underbrace{
                \begin{bNiceMatrix}
                    \mu_1\bm L & \bm O       & -\mu_1\bm I & \bm O       & \bm O       \\
                    \mu_2\bm R & \bm O       & \bm O       & -\mu_2\bm I & \bm O       \\
                    \bm O      & \mu_3 \bm D & \bm O       & \bm O       & -\mu_3\bm I
                \end{bNiceMatrix}}}
        \underset{= \bm w}{
            \underbrace{
                \begin{bNiceMatrix}
                    \bm x      \\
                    \bm \sigma \\
                    \bm u      \\
                    \bm v      \\
                    \bm \eta
                \end{bNiceMatrix}}} = \bm 0.
    }
    \label{eq:re2-aug-lop-l2l1-reg}
\end{align}
In this expression, $\bm H \in \C^{J' \times N'}, G : \C^{N'} \to \R$, and the problem can be formulated as:
\begin{equation}
    \min_{\bm w \in \C^{N'}} G(\bm w) \; \text{s.t.} \; \bm{Hw} = \bm{0}.
    \label{eq:linear-optimazation}
\end{equation}
To solve this problem, we can use the Loris Verhoeven iteration, which is one of the primal-dual methods.
The Loris Verhoeven iteration is described in \zcref{alg:loris-verhoeven}.
\begin{algorithm}[tb]
    \caption{Loris-Verhoeven iteration \cite{lorisGeneralizationIterativeSoftthresholding2011}}
    \zlabel{alg:loris-verhoeven}
    \begin{algorithmic}[line]
        \Require{$\tau_1 ,\tau_2 > 0, \bm w^0 \in \C^{N'},\bm r^0 \in \C^{J'}$}
        \Ensure{$\bm w \in \C^{N'}$}
        \For{$i = 0,1,2,\dots$}
        \State{$\tilde{\bm w}^{i+1} = \bm w^{i} + \tau_1 \bm H^*(\bm r^{i} - \tau_2\bm{Hw}^{i})$}
        \State{$\bm w^{i+1}         = \prox_{\tau_1 G}(\tilde{\bm w}^{i+1})         $}
        \State{$\bm r^{i+1}         = \bm r^{i} - \tau_2\bm{Hw}^{i+1}$}
        \EndFor
    \end{algorithmic}
\end{algorithm}
By applying \zcref{eq:re2-aug-lop-l2l1-reg} to \zcref{alg:loris-verhoeven}, we obtain the optimization algorithm for LOP-$\ell_2$/$\ell_1$ALT penalty, as shown in \zcref{alg:aug-lop-l2l1-reg}.
\begin{algorithm}[tb]
    \caption{Solver for LOP-$\ell_2$/$\ell_1$ALT \zcref{eq:aug-lop-l2l1-reg}}
    \zlabel{alg:aug-lop-l2l1-reg}
    \begin{algorithmic}[line]
        \Require{$
                \tau_1 ,\tau_2 ,\mu_1, \mu_2, \mu_3 > 0,\;
                \bm x^0 \in \C^N,\; \bm\sigma^0 \in \R^K,\;
                \bm u^0 \in \C^J,\; \bm v^0 \in \C^K,\; \bm \eta^0 \in \R^{K-1},\;
                \bm r_1^0 \in \C^J,\; \bm r_2^0 \in \C^K,\; \bm r_3^0 \in \R^{K-1},\;
                \Delta\bm r_1^0 \in \C^J,\; \Delta\bm r_2^0 \in \C^K,\; \Delta\bm r_3^0 \in \R^{K-1}
            $}
        \Ensure{$\bm x \in \C^N,\; \bm\sigma \in \R^K,\;$}
        \For{$i = 0,1,2,\dots$}
        \State{$
            \tilde{\bm x}^{i+1} = \bm x^i
            + \tau_1\mu_1\bm L^*(\bm r_1^i- \tau_2\Delta \bm r_1^{i})
            + \tau_1\mu_2\bm R^*(\bm r_2^i - \tau_2\Delta\bm r_2^{i})
        $}
        \State{$\tilde{\bm \sigma}^{i+1} = \bm\sigma^i + \tau_1\mu_3\bm D^\top (\bm r_3^{i} - \tau_2\Delta\bm r_3^{i})$}
        \State{$\tilde{\bm u}^{i+1} = \bm u^i + \tau_1\mu_1(\bm r_1^{i} - \tau_2\Delta \bm r_1^{i})$}
        \State{$\tilde{\bm v}^{i+1} = \bm v^i + \tau_1\mu_2(\bm r_2^{i} - \tau_2\Delta \bm r_2^{i})$}
        \State{$\tilde{\bm \eta}^{i+1} = \bm\eta^i + \tau_1\mu_3(\bm r_3^{i} - \tau_2\Delta \bm r_3^{i})$}
        \State{$\bm x^{i+1} = \tilde{\bm x}^{i+1}$}
        \State{$\bm u^{i+1} = \prox_{\tau_1 f}(\tilde{\bm u}^{i+1})$}
        \State{$(\bm v^{i+1}, \bm \sigma^{i+1}) = \prox_{\tau_1\lambda\varphi}(\tilde{\bm v}^{i+1}, \tilde{\bm \sigma}^{i+1})$}
        \State{$\bm \eta^{i+1} = \prox_{\iota_{B_1^\alpha}}(\tilde{\bm \eta}^{i+1}) = P_{B_1^\alpha}(\tilde{\bm\eta}^{i+1})$} by \zcref{eq:l1-ball-projection}
        \State{$\Delta \bm r_1^{i+1} = \mu_1(\bm{Lx}^i - \bm u^i)$}
        \State{$\Delta \bm r_2^{i+1} = \mu_2(\bm{Rx}^i - \bm v^i)$}
        \State{$\Delta \bm r_3^{i+1} = \mu_3(\bm{D\sigma}^i - \bm \eta^i)$}
        \State{$\bm r_1^{i+1} = \bm r_1^i - \tau_2\Delta\bm r_1^{i+1}$}
        \State{$\bm r_2^{i+1} = \bm r_2^i - \tau_2\Delta\bm r_2^{i+1}$}
        \State{$\bm r_3^{i+1} = \bm r_3^i - \tau_2\Delta\bm r_3^{i+1}$}
        \EndFor
    \end{algorithmic}
\end{algorithm}
The symbols used in \zcref{alg:aug-lop-l2l1-reg} are explained as follows.
The projection onto the $\ell_1$-norm ball with radius $\alpha$ is defined as:
\begin{equation}
    P_{B_1^{\alpha}}(\bm\eta) \coloneqq \begin{cases}
        \bm\eta,                                & \mathrm{if}\ \norm{\bm\eta}_1 \leq \alpha\,; \\
        (a_k\mathrm{sign}(\eta_k))_{k=1}^{K-1}, & \text{otherwise}.
    \end{cases}
    \label{eq:l1-ball-projection}
\end{equation}
where the coefficients $a_k$ are defined as:
\begin{align}
    a_k & \coloneqq \max\qty{\abs{\eta_k} - \frac 1 T\qty(\sum_{t=1}^T \rho_t - \alpha), 0},                   \\
    T   & \coloneqq \max\Set*{t \in \set{1, \dots, K-1}}{\frac 1t\qty(\sum_{n=1}^t \rho_n - \alpha) < \rho_t},
\end{align}
and $\rho_1$, $\dots$, $\rho_{K-1}$ are sorted in descending order of $\abs{\eta_1}$, $\dots$, $\abs{\eta_{K-1}}$.

Additionally, we evaluated the conditions under which \zcref{alg:aug-lop-l2l1-reg} converges to the optimal solution.
By performing this evaluation rigorously, we can obtain upper bounds for $\mu_1, \mu_2, \mu_3 > 0$, which can accelerate the convergence to the optimal solution of \zcref{alg:aug-lop-l2l1-reg}.
It is known that the Loris-Verhoeven iteration converges to the optimal solution when $\tau_1\tau_2\norm{\bm H}_{\mathrm{op}}^2 \leq 1$.
By applying this condition to the proposed method, $\norm{\bm H}_{\mathrm{op}}$ can be calculated as follows:
\begin{align}
     & \norm{\bm H}_{\mathrm{op}} = \sqrt{\rho(\bm H\bm H^*)} \\
     & = \small\sqrt{\rho\qty(
        \begin{bNiceMatrix}
            \mu_1^2(\bm{L}\bm{L}^* + \bm I) & \mu_1\mu_2\bm{L}\bm{R}^*        & \bm O                              \\
            \mu_1\mu_2\bm{R}\bm{L}^*        & \mu_2^2(\bm{R}\bm{R}^* + \bm I) & \bm O                              \\
            \bm O                           & \bm O                           & \mu_3^2(\bm{D}\bm{D}^\top + \bm I) \\
        \end{bNiceMatrix})},
    \label{eq:cond-mus}
\end{align}
where $\rho(\bm A)$ is the maximum eigenvalue of $\bm A$.
$\bm H\bm H^*$ can be rewritten as follows because it is a block diagonal matrix when considering the $2\times2$ and $1\times1$ block matrices separately:
\begin{subequations}
    \small
    \begin{empheq}[left=\empheqlbrace]{align}
        \rho\qty(
        \begin{bNiceMatrix}
                \mu_1^2(\bm{L}\bm{L}^* + \bm I) & \mu_1\mu_2\bm{L}\bm{R}^*        \\
                \mu_1\mu_2\bm{R}\bm{L}^*        & \mu_2^2(\bm{R}\bm{R}^* + \bm I)
            \end{bNiceMatrix}
        ) \leq (\tau_1\tau_2)^{-1}
        \label{eq:cond-mu12}\noeqref{eq:cond-mu12} \\
        \mu_3^2\qty(\norm{\bm D}_{\mathrm{op}}^2 + 1)
        \leq (\tau_1\tau_2)^{-1}.
        \label{eq:re-cond-mu3}\noeqref{eq:re-cond-mu3}
    \end{empheq}
\end{subequations}
\zcref{eq:re-cond-mu3} gives an upper bound for $\mu_3$.
On the other hand, it is difficult to explicitly solve \zcref{eq:cond-mu12} for $\mu_1, \mu_2$ unless the left-hand side matrix is block diagonal, i.e., $\bm L\bm R^* = \bm O$, and it is not possible to give upper bounds for them.
However, in the range of experiments we conducted, we were able to satisfy \zcref{eq:cond-mu12} by setting $\mu_1, \mu_2$ as follows:
\begin{subequations}
    \begin{empheq}[left=\empheqlbrace]{align}
        \displaystyle & 0 \leq \mu_1 \leq \frac{(\tau_1\tau_2)^{-1/2}}{\sqrt{2\norm{\bm L}_{\mathrm{op}}^2 + 1}}
        \\
        \displaystyle & 0 \leq \mu_2 \leq \frac{(\tau_1\tau_2)^{-1/2}}{\sqrt{2\norm{\bm R}_{\mathrm{op}}^2 + 1}}
    \end{empheq}
\end{subequations}
Therefore, although we cannot give upper bounds for $\mu_1, \mu_2$, we believe that this is not a problem in practice.

\section{Numerical experiments}
\zlabel{sec:experiments}
In this section, we demonstrate the effectiveness of the proposed method by comparing the results of denoising with TV regularization\cite{rudinNonlinearTotalVariation1992}.
However, we do not compare it with LOP-$\ell_2$/$\ell_1$ because it cannot be applied to signals with block sparse derivatives.
First, we verify the effectiveness of the proposed method on artificial data with block sparse derivatives.
Additionally, we demonstrate that over-smoothing is less likely to occur for strong noise settings, as in TV regularization, by denoising actual ion currents.

\textbf{Synthetic Examples}:
To investigate the performance of the proposed method, we performed denoising experiments on artificial data with block sparse derivatives.
The artificial data was generated using the signal $\bm x \in \mathbb{R}^{1000}$ with the same block sparsity as \zcref{fig:block-sparsity-under-linear-transform} as follows: $\bm y = \bm{x} + \bm{\epsilon}$, where $\bm y \in \mathbb{R}^{1000}$ is the observed signal and $\bm{\epsilon} \in \mathbb{R}^{1000}$ is the white Gaussian noise.
The loss function is the squared error with a regularization term.
The signal $\bm x$ is estimated from the noisy observed signal $\bm y$ under the above settings.
The evaluation index of the denoising ability is the average SNR ($= 10\log_{10}\norm{\bm x}_2^2 / \norm{\bm x - \hat{\bm x}}_2^2$) of the estimated signals $\hat{\bm x}$ obtained by performing the estimation independently 20 times.
The SNR of the signals estimated by the proposed method and TV regularization is shown in \zcref{tab:comparison-of-reconstruction-results-on-artificial-data}.
The proposed method shows better SNR than TV regularization, indicating that it is superior in denoising.
\begin{table}[tb]
    \centering
    \caption{
        Comparison of SNR of denoised signals of artificial data (\zcref{fig:block-sparsity-under-linear-transform}) with noise added to obtain the specified SNR.
        The hyperparameters are set to the values that gave the best results for each SNR.
    }
    \zlabel{tab:comparison-of-reconstruction-results-on-artificial-data}
    \begin{tabular}{c|cc}
        \hline
        noise level (SNR) & LOP-$\ell_2$/$\ell_1$ALT & TV    \\
        \hline
        10                & \textbf{25.68}           & 24.15 \\
        15                & \textbf{28.68}           & 27.54 \\
        20                & \textbf{31.83}           & 31.00 \\
        25                & \textbf{35.34}           & 34.50 \\
        30                & \textbf{38.94}           & 38.19 \\
        \hline
    \end{tabular}
\end{table}

\textbf{Application to ion current data}:
To demonstrate the robustness of the proposed method to over-smoothing, we perform denoising experiments on ion currents.
The noisy observed signal $\bm y \in \mathbb{R}^{750}$ is estimated from the signal $\bm x \in \mathbb{R}^{750}$ under the same settings as in the artificial data experiment.
The results of denoising are shown in \zcref{fig:comparison-of-reconstruction-results-on-nanopore-signal}, and the blocks estimated by the proposed method are shown in \zcref{fig:block-structure-estimation}.
From the results, it can be seen that the proposed method is less likely to be over-smoothed by large $\lambda$, in this case $\lambda \in \set{5, 10}$, compared to TV regularization.
This can be seen from the fact that over-smoothing does not occur in the proposed method when $\alpha = 0.1$, which correctly estimates the block structure, compared to the other methods.

\begin{figure}[tb]
    \centering
    \begin{tikzpicture}
        \node (img_true) {\includegraphics[width=0.97\columnwidth, keepaspectratio]{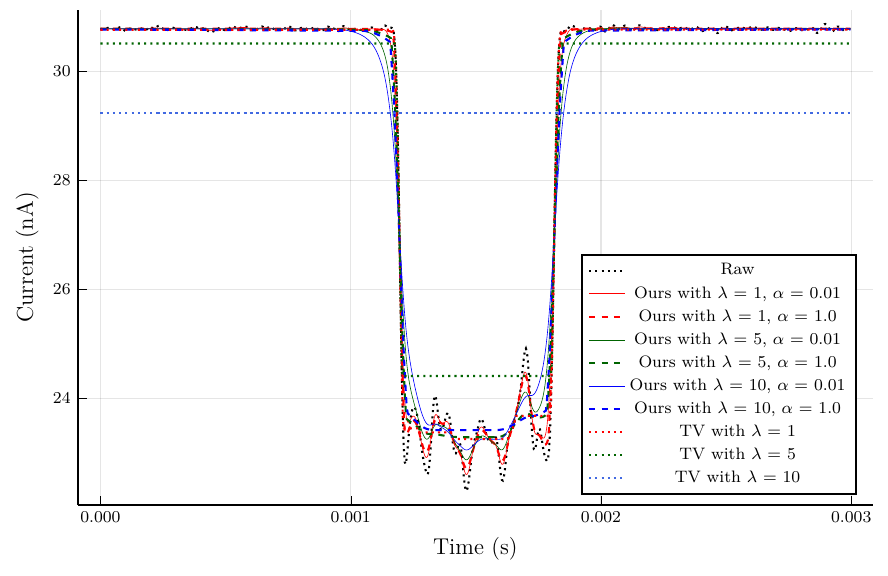}};
        \node (smoothing) at (-2.10, -1.3) {\scriptsize{\textcolor{red}{TV is smoothing}}};
        \node (empty_smoothing) at (-0.35, -1.3) {};
        \draw[->,thick,red] (smoothing) -- (empty_smoothing);
        \node (over_smoothing) at (-2.10, -0.9) {\scriptsize{\textcolor{blue}{TV is over-smoothing}}};
        \node (empty_over_smoothing_5) at (-0.35, -0.9) {};
        \node (empty_over_smoothing_10) [above=2.4cm] at (over_smoothing.center) {};
        \draw[->,thick,blue] (over_smoothing) -- (empty_over_smoothing_5);
        \draw[->,thick,blue] (over_smoothing) -- (empty_over_smoothing_10);
    \end{tikzpicture}
    \caption{
        Comparison of denoising results for ion currents.
        The red, green, and blue lines represent the estimated values for $\lambda = 1, 5, 10$, respectively.
    }
    \zlabel{fig:comparison-of-reconstruction-results-on-nanopore-signal}
\end{figure}
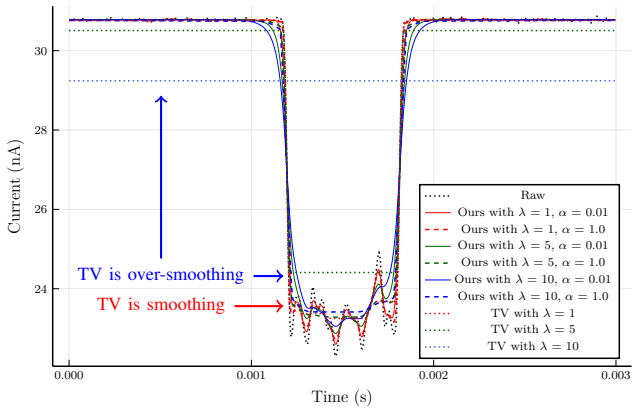

\begin{figure}[tb]
    \centering
    \includegraphics[width=\columnwidth, keepaspectratio]{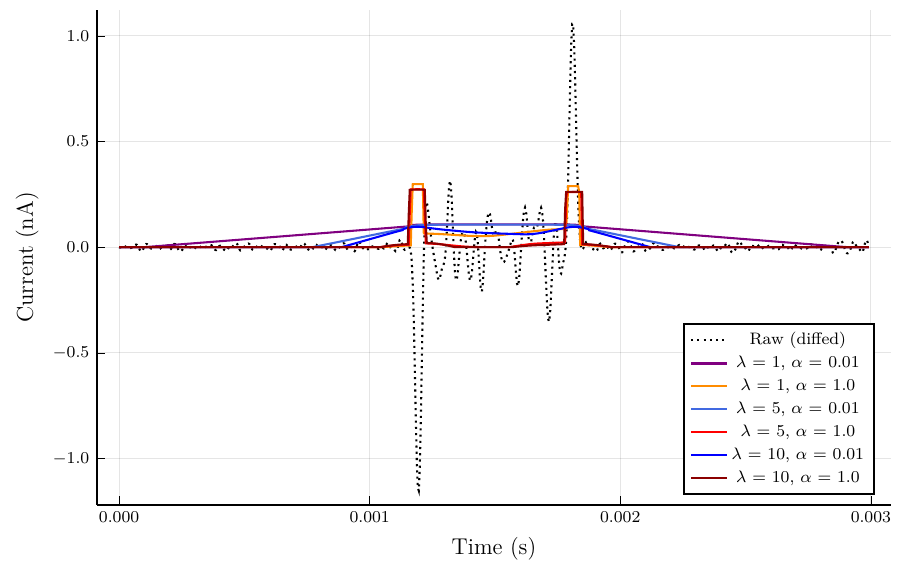}
    \caption{
        Comparison of block structure of ion current derivatives estimated by the proposed method.
    }
    \zlabel{fig:block-structure-estimation}
\end{figure}

\textbf{Application to natural image}:
We demonstrate the high denoising performance of the proposed method for removing salt-and-pepper noise in images and the effect of estimating block structure on the recovery results.
The latter result allows us to obtain a policy for selecting the regularization parameter $\lambda$ of the proposed method based on the appropriate one in TV regularization.
Since there are already many studies on selecting the appropriate parameter for TV regularization \cite{lohvitheeParameterSelectionLimited2017, meadChiTestTotal2020, afkhamLearningRegularizationParameters2021}, we can reduce the risk of choosing the wrong one for the proposed method by using them.
For this purpose, we compare the results of denoising a 256 $\times$ 256 image with the white Gaussian noise (\zcref{fig:comparison-of-reconstruction-results-on-natural-image}, right in the first row) using TV regularization and the proposed method.
The second row of \zcref{fig:comparison-of-reconstruction-results-on-natural-image} is a case where the smoothing by TV regularization is insufficient ($\lambda = 0.15$).
In this case, it is clear that the proposed method improves denoising by estimating the block structure to suppress the salt-and-pepper noise.
On the other hand, the third row of \zcref{fig:comparison-of-reconstruction-results-on-natural-image} is a case where TV regularization performs sufficient smoothing ($\lambda = 0.22$).
In contrast to the previous case, the proposed method shows lower SNR than TV regularization and blurs the image.
Therefore, it can be seen that the estimated block structure is hindering denoising.

From \zcref{fig:comparison-of-reconstruction-results-on-natural-image} results, it is expected that the regularization parameter $\lambda$ of the proposed method should be set smaller than that for TV regularization to obtain sufficient denoising.

\begin{figure}[tb]
    \centering
    \begin{tikzpicture}
        \node (img_true) {\includegraphics[width=0.48\columnwidth, keepaspectratio]{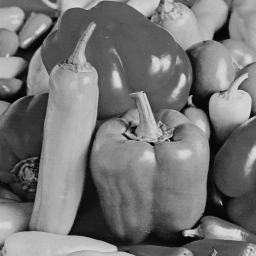}};
        \node(text_true) [below=-3pt] at (img_true.south) {Ground truth};

        \node (img_noised) [anchor=west] at (img_true.east) {\includegraphics[width=0.48\columnwidth, keepaspectratio]{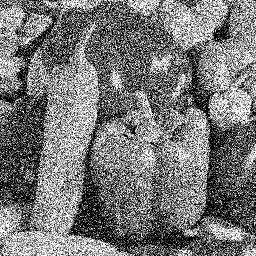}};
        \node(text_noised) [below=-3pt] at (img_noised.south) {Noisy image (5db)};

        \node (img_denoised_tv) [anchor=north] at (text_true.south) {\includegraphics[width=0.48\columnwidth, keepaspectratio]{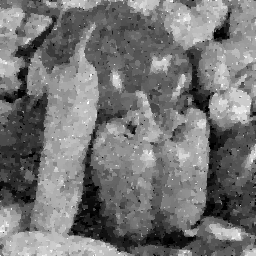}};
        \node(text_denoised_tv) [below=-3pt, align=center] at (img_denoised_tv.south) {TV Regularization \\ (16.37db, $\lambda = 0.15$)};

        \node (img_denoised_proposed) [below=-2pt, anchor=north] at (text_noised.south) {\includegraphics[width=0.48\columnwidth, keepaspectratio]{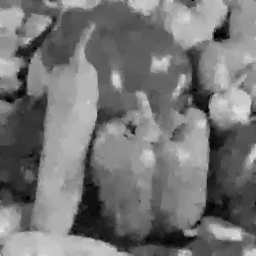}};
        \node(text_denoised_proposed) [below=-3pt, align=center] at (img_denoised_proposed.south) {Proposed Method \\ (\textbf{18.59db}, $\lambda = 0.15, \alpha = 1000$)};

        \node (img_denoised_tv_2) [anchor=north] at (text_denoised_tv.south) {\includegraphics[width=0.48\columnwidth, keepaspectratio]{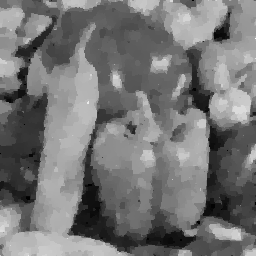}};
        \node(text_denoised_tv_2) [below=-3pt, align=center] at (img_denoised_tv_2.south) {TV Regularization \\ (\textbf{18.03db}, $\lambda = 0.22$)};

        \node (img_denoised_proposed_2) [below=-0.2pt, anchor=north] at (text_denoised_proposed.south) {\includegraphics[width=0.48\columnwidth, keepaspectratio]{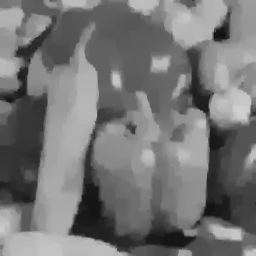}};
        \node(text_denoised_proposed_2) [below=-3pt, align=center] at (img_denoised_proposed_2.south) {Proposed Method \\ ({17.67db}, $\lambda = 0.22, \alpha = 1000$)};
    \end{tikzpicture}
    \caption{
        Comparison of denoising results for natural images.
    }
    \zlabel{fig:comparison-of-reconstruction-results-on-natural-image}
\end{figure}
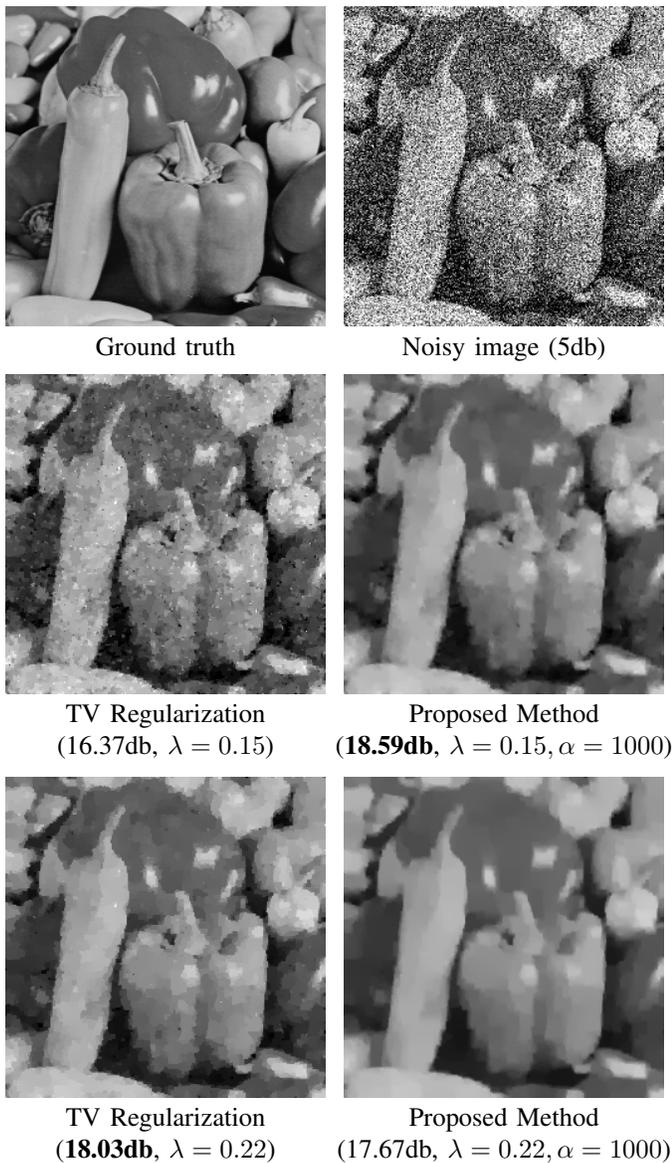

\section{Conclusion}
\zlabel{sec:conclusion}
In this paper, we proposed LOP-$\ell_2$/$\ell_1$ALT regularization and its optimization algorithm.
We also derived the conditions under which the optimization algorithm converges to the optimal solution.
The proposed method is a generalization of LOP-$\ell_2$/$\ell_1$\cite{9729560} based on non-invertible transform.
Additionally, we performed denoising experiments using LOP-$\ell_2$/$\ell_1$ALT regularization.
The proposed method is less likely to be over-smoothed than TV regularization and is superior in denoising.
As future work, it is necessary to analyze the convergence conditions of the proposed method more rigorously, demonstrate the effectiveness of block sparse regularization using non-derivative non-invertible transforms, and clarify the policy for selecting hyperparameters ($\lambda, \alpha$) appropriately.

\vspace*{1em}
\textbf{Acknowledgments}:
We thank Prof.\,Uemura and Mr.\,Akita at the University of Tokyo for providing us with ion current data from the nanopore device.

\bibliographystyle{IEEEbib}
\bibliography{myrefs}

\end{document}